%% file: main.tex
\documentclass[letterpaper]{article} 
\usepackage{aaai2026}  
\usepackage{times}  
\usepackage{helvet}  
\usepackage{courier}  
\usepackage[hyphens]{url}  
\usepackage{graphicx} 
\usepackage{natbib}  
\usepackage{caption} 
\usepackage{algorithm}
\usepackage{algorithmic}

\usepackage{amsmath} 
\usepackage{enumitem}
\newcommand{\M}{{Owl}}
\usepackage{amsmath,bm,amssymb}
\usepackage{booktabs}
\usepackage{multirow}
\usepackage{svg}
\usepackage{newfloat}
\usepackage{listings}
\DeclareCaptionStyle{ruled}{labelfont=normalfont,labelsep=colon,strut=off} 
\floatstyle{ruled}
\newfloat{listing}{tb}{lst}{}
\floatname{listing}{Listing}
\title{
Causally-Grounded Dual-Path Attention Intervention for Object Hallucination Mitigation in LVLMs
}
\author{
    Liu Yu\textsuperscript{\rm 1,2}\thanks{This work was conducted during Liu Yu’s joint PhD program at the University of Auckland, thanks to the support of the China Scholarship Council (CSC, Grant No. 202506070076).},
    Zhonghao Chen\textsuperscript{\rm 1},
    Ping Kuang\textsuperscript{\rm 1}\thanks{Corresponding author.},
    Zhikun Feng\textsuperscript{\rm 1},
    Fan Zhou\textsuperscript{\rm 1},
    Lan Wang\textsuperscript{\rm 1},
    Gillian Dobbie\textsuperscript{\rm 2}
}
\affiliations{
    \textsuperscript{\rm 1}University of Electronic Science and Technology of China,
    \textsuperscript{\rm 2}University of Auckland\\
liu.yu@std.uestc.edu.cn,
202321090211@std.uestc.edu.cn,
    kuangping@uestc.edu.cn,
    202411090917@uestc.edu.cn,
    fan.zhou@uestc.edu.cn,
    202421090210@std.uestc.edu.cn,
    g.dobbie@auckland.ac.nz
}

\usepackage{bibentry}

\begin{document}

\maketitle

\begin{abstract}
Object hallucination remains a critical challenge in Large Vision-Language Models (LVLMs), where models generate content inconsistent with visual inputs. Existing language-decoder based mitigation approaches often regulate visual or textual attention independently, overlooking their interaction as two key causal factors. 
To address this, we propose \textbf{\M}
(Bi-m\underline{\textbf{O}}dal attention re\underline{\textbf{W}}eighting for \underline{\textbf{L}}ayer-wise hallucination mitigation),
a causally-grounded framework that models hallucination process via a structural causal graph, treating decomposed visual and textual attentions as mediators.
We introduce VTACR (Visual-to-Textual Attention Contribution Ratio), a novel metric that quantifies the modality contribution imbalance during decoding. Our analysis reveals that hallucinations frequently occur in low-VTACR scenarios, where textual priors dominate and visual grounding is weakened. To mitigate this, we design a fine-grained attention intervention mechanism that dynamically adjusts token- and layer-wise attention guided by VTACR signals.
Finally, we propose a dual-path contrastive decoding strategy: one path emphasizes visually grounded predictions, while the other amplifies hallucinated ones -- letting visual truth shine and hallucination collapse. 
Experimental results on the POPE and CHAIR benchmarks show that \M~achieves significant hallucination reduction, setting a new SOTA in faithfulness while preserving vision-language understanding capability.
Our code is available at 
\url{https://github.com/CikZ2023/OWL}
\end{abstract}


\input{secs/1-intro}

\input{secs/2-related_work}
\input{secs/3-preliminary}

\input{secs/4-Causal_interventions}

\input{secs/5-method}

\input{secs/6-experiments}
\input{secs/7-conclusion}

\bibliography{main}
\newpage
\clearpage 

\end{document}

%% file: secs/1-intro.tex
\section{Introduction}

\begin{figure}[t]
\centering
\includegraphics[width=1\linewidth]{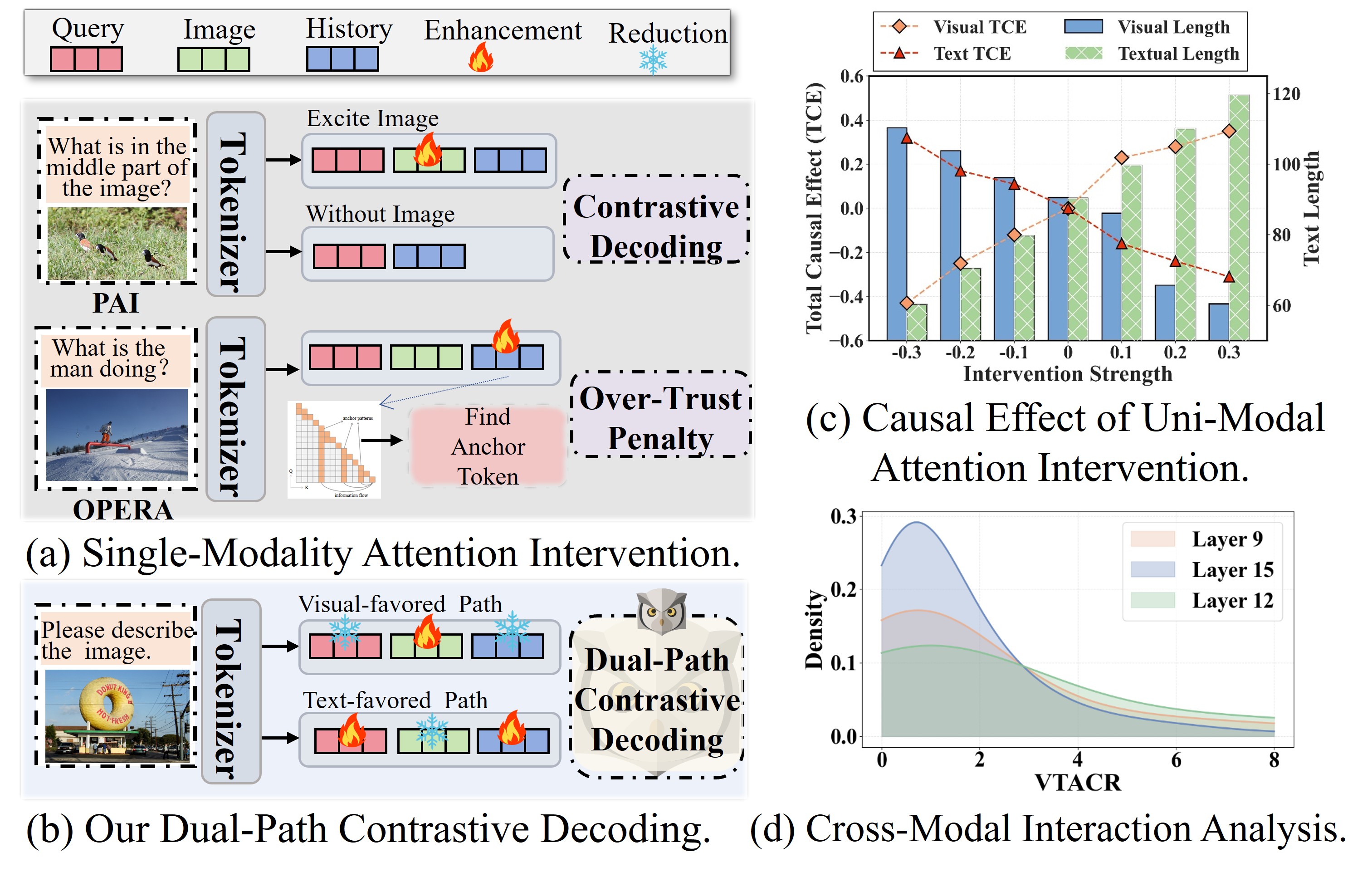}
\caption{
Motivation of our work. 
(a) Existing methods manipulate attention in a single modality (visual or text).
(b) We contrast the visual-favored path and text-favored path based on the VTACR-guided attention calibration.
(c) Increasing visual attention improves causal effect but shortens output, while increasing textual attention has the opposite impact.
(d) Hallucinated tokens typically show lower VTACR, indicating a skewed visual-to-textual modality reliance.
}
\label{fig:intro}
\vspace{-3mm}
\end{figure}

Large Vision-Language Models (LVLMs) such as MiniGPT-4~\cite{zhu2023minigpt}, LLaVA~\cite{liu2023visual} and Shikra~\cite{chen2023shikra}, have achieved impressive progress in image-based text generation~\cite{li2023towards,li2022contrastive}, allowing a wide range of applications, from visual question answering~\cite{kim2025visual} to open-ended image description~\cite{liu2024mmdu,yu2025bimodal}.
Despite their success, these models remain vulnerable to a persistent issue -- \textit{object hallucination} -- that generates mentions of objects not present in the image.
Such hallucinations not only undermine the trustworthiness of LVLMs but also pose serious risks in safety-critical domains such as medical imaging~\cite{hu2024omnimedvqa},  robotic navigation~\cite{lange2025general}, etc.

To mitigate this issue, existing approaches span several directions. Early efforts align LVLMs with human preferences~\cite{LLaVA-RLHF,gunjal2024detecting} through reinforcement learning or feedback-based fine-tuning, which improves consistency but often requires costly annotations. Others adopt post-processing strategies~\cite{LURE,CGD,woodpecker} using external modules to detect or revise hallucinated entities after generation. More recently, decoding optimizations have gained traction: Some~\cite{VCD} perturb visual inputs to reveal unstable predictions, while others~\cite{liu2407paying,huang2024opera} manipulate attention weights to boost visual grounding or suppress over-reliance on previous tokens.
However, as shown in Figure~\ref{fig:intro}(a), language-decoder attention-based methods tend to act on either the visual path -- by enhancing attention to image tokens -- or the textual path -- by diminishing influence from autoregressive history. This uni-modal design overlooks the attention imbalance between the visual and textual modalities, which often lie at the heart of hallucination. 
To better understand this issue, we decompose the attentions in language-decoder into visual and textual aspects, and analyze both uni-modal and their interaction. As shown in Figure~\ref{fig:intro}(c), we observe that: (1) solely enhancing visual attention consistently reduces hallucinations (indicated by increasing Total Causal Effect, TCE) but leads to shorter outputs; (2) in contrast, solely increasing textual attention (including query and history) expands output length but aggravates hallucinations. This tradeoff suggests the need to balance their reliance rather than treating them independently.
To measure their interaction, we introduce a new metric -- visual-to-textual attention contribution
ratio (VTACR) -- to quantify the relative hallucination contribution of visual versus textual signals for the current token during generation:
\begin{itemize}
    \item \textbf{Visual Token Attention Contribution}:
\end{itemize}
    \begin{equation}
    \nu^{(\ell)} = \frac{1}{{N}|\mathcal{V}|} \sum_{j \in \mathcal{V}} \sum_{i=1}^{N} \mathbf{A}^{(\ell)}_{i,j}
    \end{equation}
where $\mathcal{V}$ is the set of indices for visual prefix tokens, 
and \(N\) is the number of attention heads. $\mathbf{A}^{(\ell)}_{i,j}$ indicates 
the visual attention weight of the current token in the $i$-th head and the $\ell$-th layer, and $\nu^{(\ell)}$ is the average of the attention weight of the current token to the visual prefix tokens.
\begin{itemize}
    \item \textbf{Text Token Attention Contribution}:
\end{itemize}
    \begin{equation}
    \tau^{(\ell)} = \frac{1}{{N}|\mathcal{T}|} \sum_{k \in \mathcal{T}} \sum_{i=1}^{N} \mathbf{A}^{(\ell)}_{i,k}
    \end{equation}
where $\mathcal{T}$ is indices set for query and history text prefix tokens, $\mathbf{A}^{(\ell)}_{i,k}$ indicates the text attention weight of the current token in  $i$-th head, $\ell$-th layer, and $\nu^{(\ell)}$ is the average of the attention weight of the current token to the text prefix tokens.
\begin{itemize}
    \item \textbf{Layer-wise VTACR}:
\end{itemize}
    \begin{equation}
    \text{VTACR}^{(\ell)} = \frac{\nu^{(\ell)}}{\tau^{(\ell)}}\label{equa:V}
    \end{equation}
This ratio measures the relative contribution of visual tokens to text tokens in the $\ell$-th layer. $L$ is the total layers. 
As shown in Figure~\ref{fig:intro}(d), hallucinated tokens tend to exhibit skewed VTACR values in LLaVA-1.5, revealing a tendency to over-rely on textual modality while neglecting visual grounding (consistent across layers/backbones).
This modality imbalance motivates us to explicit decomposition of two contrasting attention pathways: a \textit{vision-favored} path that reinforces grounded reasoning, and a \textit{text-favored} path that tends to preserve hallucinated content. Such separation enables us to capture the asymmetric roles of each modality in hallucination formation and lays the foundation for a contrastive decoding mechanism.

To formalize this intuition, we construct a Structural Causal Model (SCM) in which visual and textual attention serve as \textit{mediators} between inputs and outputs. Unlike traditional causal interventions on inputs or latent states~\cite{huang2024brings,CausalMM}, mediator-based intervention enables direct manipulation of attention weights while preserving input consistency. This provides a more interpretable and fine-grained view of how modality interactions drive hallucinations.
We further leverage VTACR to guide token- and layer-wise attention adjustment. Rather than applying fixed scaling factors (as in previous work like PAI~\cite{liu2407paying}), by adjusting the token-by-token attention weights according to real-time modality contributions, especially in layers with pronounced asymmetry -- our \M~not only corrects attention imbalance, but also improves generation quality as a side benefit, producing longer outputs with fewer repetitions and less overcorrection.
Finally, we introduce a dual-path contrastive decoding strategy, incorporating both vision- and text-favored decoding paths. The contrast amplifies the distinction between hallucinated and faithful tokens, effectively mitigating hallucinations.
Our contributions are four-fold: \begin{itemize}[leftmargin=*,itemsep=0pt,topsep=0pt,parsep=0pt]
    \item We introduce a novel metric VTACR to quantify cross-modal reliance during generation, and use it to guide fine-grained token- and layer-wise attention modulation.
    \item We formulate a SCM where visual and textual attention serve as mediators, enabling interpretable dual-modality interventions to analyze the hallucinations process.
    \item We propose a VTACR-guided dual-path contrastive decoding strategy that exaggerates modality bias -- amplifying faithful generations while exposing hallucinated ones -- thereby enabling effective suppression through contrast.
    \item 
    Experiments on POPE/CHAIR show \M~achieves notable hallucination reduction, with 22.9\% improvement on CHAIR, while evaluations on five VQA benchmarks confirm preserved vision-language understanding capability.
\end{itemize}

%% file: secs/2-related_work.tex
\section{Related Work} 
\paragraph{Object Hallucination Mitigation in LVLMs.}
Object hallucination in LVLMs often stems from over-reliance on spurious correlations -- models tend to exploit shortcut patterns, such as object co-occurrence or prompt biases, learned from large-scale vision-language data. This results in fluent but visually unfaithful generations~\cite{lyu2025existingtestingtoolsreally,zhou2024association,ICAS}.
To address this, existing methods fall into three main paradigms:
(1) Early \textit{human preference alignment} approaches like LLaVA-RLHF~\cite{LLaVA-RLHF} and  instruction tuning fine-tune models to match human-preferred responses, ~\cite{bai2025mitigating}. While helpful for fluency and helpfulness, these methods are costly and offer limited interpretability into hallucination origins.
(2) \textit{Post-processing} works like LURE~\cite{LURE}, CGD~\cite{CGD}, and Woodpecker~\cite{woodpecker} detect hallucinations post-hoc via confidence scoring or visual grounding checks.
These modular approaches offer flexibility but rely heavily on external cues and do not address the root cause.
(3) Recent \textit{decoding optimization} works proactively steer generation to reduce hallucinations. Contrastive decoding methods (e.g., VCD~\cite{VCD},
HIO~\cite{HIO}) amplify hallucinated signals to isolate faithful ones. Others manipulate attention to strengthen visual grounding~\cite{liu2407paying} or suppress misleading text~\cite{huang2024opera}. Token-level signals like VAR~\cite{jiang2024devils} and ``attention sin'' patterns~\cite{zhang2024seeing} enhance interpretability, yet most still intervene on a single modality, overlooking the joint causal role of visual and textual attention.

\paragraph{Causality in LVLMs.}
Causal inference~\cite{pearl2010causal,neal2020introduction} offers a powerful lens to enhance interpretability and robustness in AI systems, particularly do-calculus or counterfactual simulations to uncover causal links in language generation~\cite{zhang2025causal}, visual question answering~\cite{zhang2025open}, and fairness analysis~\cite{yu2025bridging,zhou2023causal}. These approaches typically intervene on inputs or latent features to disentangle causal effects from spurious correlations. 
Recent works use this tool for object hallucination: \citet{huang2024brings} intervenes on image/text inputs and embeddings to analyze hallucination triggers, while CausalMM~\cite{CausalMM} perturbs attentions in vision and language decoder to probe modality priors. 
However, they often rely on coarse-grained manipulations or treat attention as a black box.
In contrast, we model visual and textual attention as explicit mediators within a SCM, allowing fine-grained mediator interventions that directly adjust internal attention without altering the input, enabling interpretable causal analysis of modality influence. 

%% file: secs/3-preliminary.tex
\section{Preliminary}


\noindent\textbf{Formulation of LVLM Generation.}
The LVLMs typically consist of three key components: a visual encoder, a cross-modal projector, and a language decoder. The visual encoder (e.g., ViT~\cite{vit}) extracts a sequence of image features $\mathbf{X}_V = [{x}_{v_1}, \ldots, {x}_{v_N}]$, which are mapped into the text embedding space via a cross-modal projector~\cite{alayrac2022flamingo,li2023blip,liu2023visual}. The projected visual tokens are then concatenated with the instruction text $\mathbf{X}_T = [{x}_{i_1}, \ldots, {x}_{i_M}]$ and historical textual input $\mathbf{X}_H = [{x}_{h_1}, \ldots, {x}_{h_L}]$, forming the input to the language decoder (e.g., LLaMA~\cite{touvron2023llama}).

The decoder integrates the multi-modal inputs via multi-layer attention and produces contextualized hidden states. 
The hidden state $\mathbf{h}_t$ at a target position $t$ (typically the last token in $\mathbf{X}_H$) is used to compute the token probability:
\begin{align}
    \mathbf{h}_t &= \mathrm{Decoder}_\theta(\mathbf{X}_V, \mathbf{X}_I, \mathbf{X}_H)[t], \\
    P_\theta(y_t \mid \mathbf{X}_V, \mathbf{X}_T, \mathbf{X}_H) &= \mathrm{Softmax}(\mathbf{W}_o \cdot \mathbf{h}_t),
\end{align}
where $\mathbf{W}_o \in \mathbb{R}^{|\mathcal{V}| \times d}$ is the output projection matrix and $|\mathcal{V}|$ denotes the vocabulary size. Autoregressive decoding continues until an end-of-sequence (EOS) token is produced, forming the final output $y_{1:T}$.
The goal of object hallucination mitigation is to ensure that the generated object-level content aligns faithfully with the visual evidence in $\mathbf{X}_V$, such that the outputs are both semantically relevant and visually grounded. The overall framework is shown in Figure~\ref{fig:framework}.

%% file: secs/4-Causal_interventions.tex
\section{Methodology} 

\begin{figure}[tp]
\centering
\includegraphics[width=0.55\linewidth]{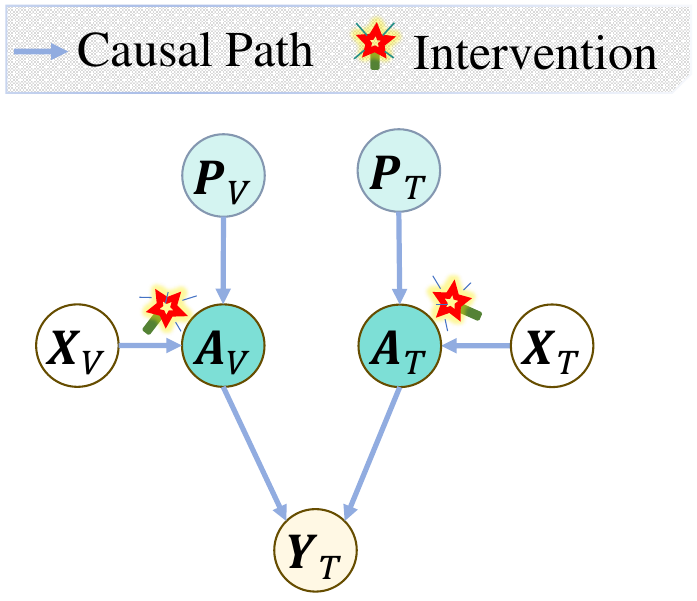}
\caption{
The SCM for analyzing the hallucination process. Visual input ($X_V$) and text input ($X_T$) affect the output ($Y_T$) via visual attention ($A_V$) and text attention ($A_T$). Visual priors ($P_V$) and language priors ($P_T$) confound the attention paths and may cause hallucinations. Interventions on $A_V$ and $A_T$ help estimate their causal impact.
}
\label{fig:scm}
\end{figure}
\begin{figure*}
\centering
\includegraphics[width=0.94\linewidth]{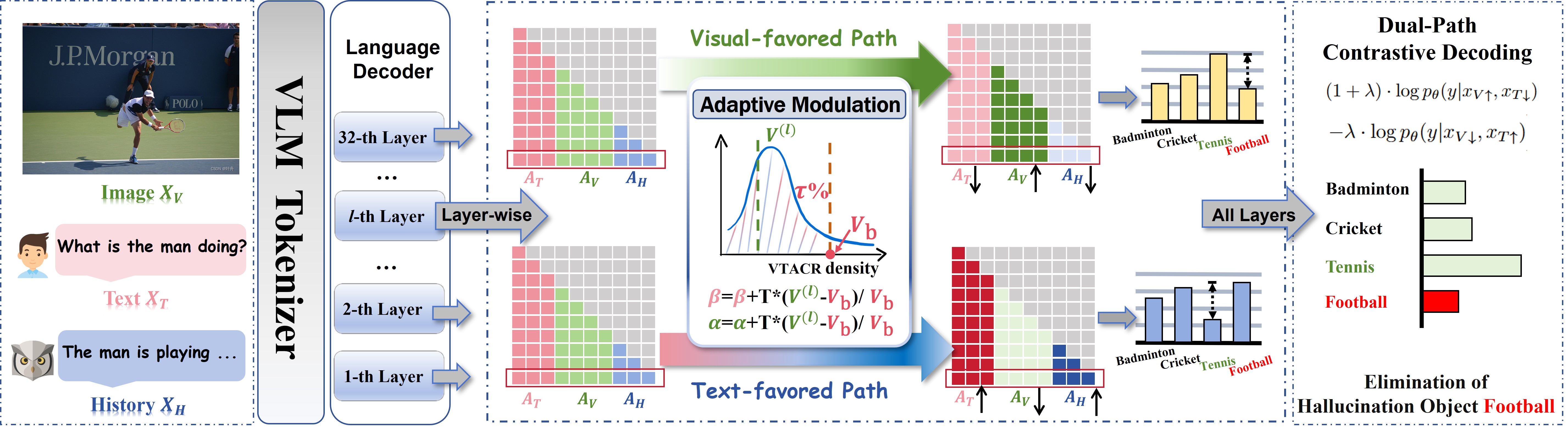}
\caption{The overall framework of \textbf{\M}. Given 
image, text, and generation history, \M~performs layer-wise decomposition of visual, textual, and historical attentions. Based on the VTACR distribution, \M~adaptively modulates attention along: a visual-favored path (enhancing grounding) and a text-favored path (amplifying hallucination). A dual-path contrastive decoding strategy then drives the LVLM to suppress hallucinations (e.g., Football) while preserving truthful predictions.}
\label{fig:framework}
\vspace{-3mm}
\end{figure*}

\paragraph{Causal Modeling of Object Hallucination Process.}
We construct a SCM in Figure~\ref{fig:scm}. 
The attention mechanisms in LLaMA-style language decoders serve as a core computational unit, and we explicitly decouple attention into visual and textual attention components, which are influenced by both input modalities and modality-specific priors.
Specifically, the SCM consists of image input $X_V$, text input $X_T$, priors $P_V$ and $P_T$, visual and textual attention $A_V$ and $A_T$, and final language output $Y_T$. The causal relations can be summarized as:
\begin{align*}
X_V \rightarrow A_V \rightarrow Y_T, \quad & P_V \rightarrow A_V \rightarrow Y_T, \\
X_T \rightarrow A_T \rightarrow Y_T, \quad & P_T \rightarrow A_T \rightarrow Y_T.
\end{align*}
where these paths reflect how hallucinations may be introduced via unbalanced attention induced by modality-specific inputs and priors. Notably, since the priors $P_V$ and $P_T$ are unobservable or non-manipulable, we cannot intervene on them directly. However, as they influence the output exclusively through the \textit{mediators} $A_V$ and $A_T$, we target these attention modules for causal intervention.
To this end, 
we apply soft interventions on $A_V$ and $A_T$:
\begin{align}
do(A_V = A_V^*), \quad do(A_T = A_T^*),
\end{align}
where $A_V^*$ and $A_T^*$ are calibrated attention weights obtained through our debiasing module. This mediator intervention conforms to the do-calculus formulation in causal inference~\cite{pearl2022direct}, representing a hypothetical manipulation that isolates the effect of attention from upstream biases.

To evaluate the effect of such interventions, we adopt the Total Causal Effect (TCE) as our primary metric. TCE measures the average change in hallucination behavior caused by modifying the mediators, and is defined as:
\begin{align}
\text{TCE} &= \mathbb{E}_{x \sim X_{\text{test}}} \big[ \Psi\big(Y_T(P, A), Y_T'(P, A^*)\big) \big]\\
\Psi\big(Y_T, Y_T'\big) &= 2 \cdot \mathbb{I} \left( H(Y_T) > H(Y_T') \right) - 1
\end{align}
where $\Psi$ measures the causal effect metric between the distributions before and after the intervention. $H(\cdot)$ denotes the hallucination evaluation benchmark CHAIR in Sec.~\ref{sec:expr}, and $Y_T'$ is the output under intervened attention. 
\(I(\cdot)\) denotes the indicator function. If \(H(Y_T) > H(Y_T')\) holds, the indicator result is $1$, indicating that the hallucination is reduced; otherwise, it is $0$, indicating that the hallucination is not reduced.

\noindent \textbf{Mediator Analysis.} Figure~\ref{fig:intro} (c) illustrates the results of causal interventions on modality-specific attention. Increasing visual attention weights \( A_V \) leads to a measurable reduction in hallucination scores \( H(Y_T) \), confirming a positive TCE from \( A_V \) to the output \( Y_T \). This indicates that enhancing \( A_V \) effectively strengthens the causal path \( X_V \rightarrow A_V \rightarrow Y_T \), thereby reducing the influence of biased visual priors \( P_V \) and promoting image-grounded reasoning. In contrast, increasing textual attention \( A_T \) results in elevated \( H(Y_T) \), suggesting that amplifying linguistic priors \( P_T \) can disrupt modality alignment and induce hallucinations. Conversely, suppressing \( A_T \) helps mitigate this effect. These findings validate \( A_V \) and \( A_T \) as mediators in the causal paths \( P_V \rightarrow A_V \rightarrow Y_T \) and \( P_T \rightarrow A_T \rightarrow Y_T \). Through soft interventions \( do(A_V = A_V^*) \), \( do(A_T = A_T^*) \), we demonstrate that attention allocation serves not only an architectural function but also a causal mechanism for controlling hallucinations in vision-language reasoning.

%% file: secs/5-Method.tex
\paragraph{Adaptive Attention Modulation.}
To enable fine-grained hallucination suppression, we leverage the proposed VTACR to adaptively apply attention re-weighting.
Specifically, we randomly sample $2,000$ hallucinated samples from MSCOCO~\cite{lin2014microsoft} and compute the VTACR value $V^{(\ell)}$ via Equation~\eqref{equa:V} for hallucinated tokens in each decoder layer $\ell$. 
By aggregating these values, we estimate a layer-wise VTACR density distribution. For each layer $\ell$, 
we define the base score $V_\text{b}^{(\ell)}$ as the distribution's $\tau$-th percentile,
where $\tau$ is a hyperparameter.
During generation, we obtain per-layer $V^{(\ell)}$ for the current token. If $V^{(\ell)} < V_\text{b}^{(\ell)}$, indicating insufficient visual grounding, we increase the visual/textual attention coefficient $\alpha$, $\beta$. Otherwise, we retain their original values. This modulation is formulated as:
\begin{align}
\tilde{T}^{(\ell)} &= \mathbb{I}(V^{(\ell)} < V_\text{b}^{(\ell)}) \cdot \min\left(T \cdot \frac{V^{(\ell)} - V_\text{b}^{(\ell)}}{V_\text{b}^{(\ell)}}, T\right), \\
\tilde{\alpha}^{(\ell)} &= \alpha + \tilde{T}^{(\ell)}, \label{equ:visual_att_weight}\\
\tilde{\beta}^{(\ell)} &= \beta + \tilde{T}^{(\ell)}, \label{equ:text_att_weight}
\end{align}
where $T$ is a pre-defined modulation coefficient, and $\mathbb{I}(\cdot)$ is an indicator function that activates intervention only when the current $V^{(\ell)} < V_\text{b}^{(\ell)}$. This mechanism allows the model to dynamically prioritize visual/text features when hallucination risk is detected, while avoiding unnecessary intervention for well-grounded tokens.
\begin{figure}[t]
\centering
\includegraphics[width=1\linewidth]{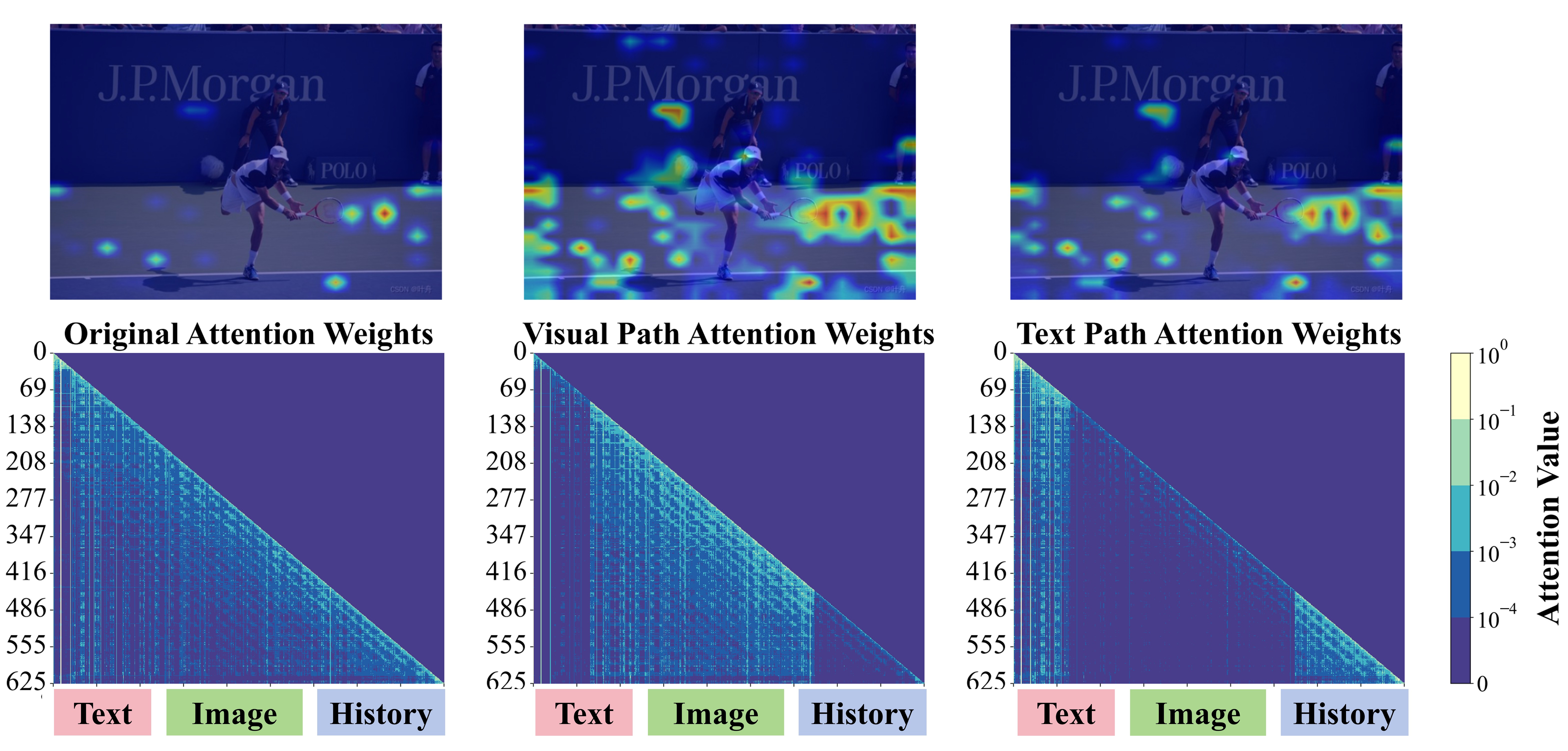}
\caption{
Visualization of dual-path attention intervention.  
Compared to original attention, the visual- and text-favored paths highlight distinct modality preferences in 
token-level.
}\vspace{-3mm}
\label{fig:attention}
\end{figure}
\paragraph{Dual-path Attention Intervention and Contrastive Decoding.}
Unlike PAI that perform uniform intervention at the attention layer, our analysis based on the VTACR signal reveals a critical insight: the influence of visual and textual attention on hallucination varies significantly across different layers. Therefore, we propose attention intervention based on the contribution of each layer.
We disentangle modality dominance by creating two hypothetical but informative paths: a \textit{visual-favored} decoding path that amplifies visual grounding, and a \textit{text-favored} path that mimics excessive textual reliance. 
This dual-path formulation enables us to assess and contrast the causal impact of $A_V$ and $A_T$, and make more targeted corrections. 

In the \textit{visual-favored path}, we enhance the attention on visual tokens $A_V$ and attenuate those on textual tokens $A_T$, forcing the model to rely more on image evidence:
\begin{align}
\tilde{\mathbf{A}}_{i,j}^{(\ell)} &= \mathbf{A}_{i,j}^{(\ell)} + \tilde{\alpha}^{(\ell)}  \cdot |\mathbf{A}_{i,j}^{(\ell)}| , \quad (j \in \mathcal{V})\label{eqa:vp_alpha}\\
\tilde{\mathbf{A}}_{i,k}^{(\ell)} &= \mathbf{A}_{i,k}^{(\ell)} - \tilde{\beta}^{(\ell)} \cdot |\mathbf{A}_{i,k}^{(\ell)}| , \quad (k \in \mathcal{T})\label{eqa:vp_beta}
\end{align}

Conversely, the \textit{text-favored path} downplays visual grounding and encourages decoding dominated by textual priors, simulating hallucination risks:
\begin{align}
\tilde{\mathbf{A}}_{i,j}^{(\ell)} &= \mathbf{A}_{i,j}^{(\ell)} - \tilde{\alpha}^{(\ell)} \cdot |\mathbf{A}_{i,j}^{(\ell)}| , \quad (j \in \mathcal{V})\label{eqa:tp_alpha}\\
\tilde{\mathbf{A}}_{i,k}^{(\ell)} &= \mathbf{A}_{i,k}^{(\ell)} + \tilde{\beta}^{(\ell)} \cdot |\mathbf{A}_{i,k}^{(\ell)}| , \quad (k \in \mathcal{T})\label{eqa:tp_beta}
\end{align}
where an example is visualized in Figure~\ref{fig:attention}. The original attention exhibits dispersed focus over both text and image tokens, leading to ambiguous grounding. Under our VTACR-guided modulation, the visual-favored path sharpens attention toward salient image regions (e.g., the player and racket), while the text-favored path highlights prior textual tokens, potentially driving hallucinations. 

Above two adaptive attention  intervention paths produce logit distributions that reflect divergent modality dependencies. 
Based on this, we propose a Dual-path Contrastive Decoding (DCD) that explicitly simulates and contrasts two complementary decoding trajectories.
By contrasting them, we can detect and suppress hallucination-prone outputs. The final prediction is obtained via contrastive fusion:
\begin{align}
P_{\text{DCD}}(Y|X_V, X_I) &= \text{Softmax} \Big[ (1+\lambda) \cdot \log p_{\theta}(y|X_{V\uparrow}, X_{T\downarrow}) \nonumber\\
&\quad - \lambda \cdot \log p_{\theta}(y|X_{V\downarrow}, X_{T\uparrow}) \Big]\label{eq:CDC}
\end{align}
where $(X_{V\uparrow}, X_{T\downarrow})$ and $(X_{V\downarrow}, X_{T\uparrow})$ denote the visual and text favored decoding settings, respectively, and $\lambda$ is a tunable contrastive strength. 
Combined with VTACR-guided attention calibration, this dual-path approach completes a causal intervention loop: measuring, adjusting, and decoding -- all guided by mediator behavior.

%% file: secs/6-experiments.tex
\section{Experiments}\label{sec:expr}

\paragraph{Benchmarks.}
We test \M~on two hallucination detection benchmarks and one comprehensive evaluation metric:

\textbf{(1) POPE}~\cite{pope}: The Polling-based Object Probing Evaluation assesses object hallucinations through a binary QA-style probing interface. It queries the model about the presence of specific objects in an image, bypassing the need for caption parsing, and providing a reliable, model-agnostic measurement.

\textbf{(2) CHAIR}~\cite{rohrbach2018object}: The Caption Hallucination Assessment with Image Relevance evaluates hallucinations at both the instance and sentence levels. CHAIR$_I$ quantifies the proportion of hallucinated objects among all mentioned objects, while CHAIR$_S$ measures the percentage of captions containing at least one hallucinated object.

\textbf{(3) GPT-4V Assisted Evaluation}: Following~\cite{huang2024opera,liu2407paying}, we adopt GPT-4V~\cite{gpt-4v} 
to comprehensively evaluate the semantic correctness and visual faithfulness of the generated captions. 

\paragraph{Models and Baselines.}
We evaluate our approach on three representative LVLM backbones, each covering a distinct architectural paradigm:
\textbf{LLaVA-1.5}~\cite{liu2024improved}: An alignment-optimized vision-language model based on CLIP and Vicuna;
\textbf{MiniGPT-4}~\cite{zhu2023minigpt}: A task-agnostic LVLM with Vicuna-based decoding and BLIP-2-style alignment;
\textbf{Shikra}~\cite{chen2023shikra}: A structured LVLM that supports object-level localization and fine-grained grounding.
For comparison, we include both classic decoding strategies -- Beam Search, Greedy Decoding, and Nucleus Sampling -- and several state-of-the-art baselines:
\textbf{VCD}~\cite{VCD}: Introduces visual contrastive decoding to suppress language bias and enhance visual grounding.
\textbf{PAI}~\cite{liu2407paying}: Modulates attention via perplexity-aware gating to improve robustness under ambiguous conditions.
\textbf{OPERA}~\cite{huang2024opera}: Prevents repetitive hallucinations through rollback and attention weight suppression.
\textbf{CausalMM}~\cite{CausalMM}: Utilizes a causal diagram to apply counterfactual reasoning to both the vision encoder and language decoder.

\input{tables/CHAIR_result}

\input{tables/POPE_results}

\paragraph{Configurations and Parameters.}
All experiments are conducted on $500$ images randomly sampled from the MSCOCO val2014 dataset~\cite{lin2014microsoft}, following the setup in prior works~\cite{huang2024opera,liu2407paying}. 
The visual/textual attention coefficient $\alpha$, $\beta$ are empirically tuned for each model to balance the quality of the generation and the reduction of hallucinations. Specifically, we set $(\alpha{=}0.4, \beta{=}0.5)$ for LLaVA-1.5, $(0.2, 0.3)$ for MiniGPT-4, and $(0.5, 0.3)$ for Shikra. The contrastive decoding strength $\lambda$ is fixed at 0.2, the modulation coefficient $T$ is set to 0.2 across all models, and the default value of $\tau$ is 80.
All experiments are conducted on $4\times$NVIDIA $3090$ GPUs. 
The reported results are the best of $20$ runs for all models and the statistical significance of the results is less than $0.05$, i.e., $p < 0.05$.

\noindent\textbf{Results on CHAIR hallucination evaluation.}

Table~\ref{tab:chair} shows
\M~consistently outperforms all baselines across three LVLMs in both sentence-level ($C_S$) and instance-level ($C_I$) hallucination metrics. Compared to the strongest prior method PAI, our \M~achieves substantial $C_S$ reductions of $17.6\%$, $14.5\%$, and $22.1\%$, and $C_I$ reductions of $21.4\%$, $36.7\%$, and $24.8\%$ on LLaVA-1.5, MiniGPT-4, and Shikra, respectively. Importantly, this hallucination suppression is achieved while preserving or even improving generation length, indicating that our intervention does not compromise output richness.
These gains stem from our adaptive, VTACR-guided dual-path decoding strategy. Unlike PAI or OPERA that apply static or uni-modal attention shifts, our approach dynamically adjusts visual and textual attention at each decoding layer based on token-level VTACR scores. This enables effective hallucination suppression without overcorrecting or truncating informative content -- striking a better balance between grounding and fluency.
Notably, though CausalMM intervenes attention in both visual encoder and LLM layers, it inherently amplifies hallucinatory signals instead of enhancing image-awareness. Unlike our DCD, which sufficiently widens the gap between faithful and hallucination tokens, this limits its efficacy.

\noindent\textbf{Results on POPE generalization benchmark.} Table~\ref{tab:pope} reports results under Random, Popular, and Adversarial settings. \M~consistently outperforms classic decoding baselines (Beam Search, Greedy, Nucleus) across all LVLMs and splits, with particularly strong gains in adversarial scenarios. 
Besides, \M~outperforms other hallucination mitigation methods in a competitive manner in most cases. While slightly trailing PAI on MiniGPT-4 and LLaVA-1.5 under the Popular setting, we speculate that this setting prioritizes high-frequency objects, which aligns better with PAI's scenario of textual inertia. 
For Shikra, \M~delivers the highest accuracy in three settings.
These improvements highlight the robustness of \M~under varied linguistic priors.
By explicitly contrasting visually grounded and hallucination-prone paths, \M~enhances model generalization.

\begin{figure}[t]
\centering
\includegraphics[width=1\linewidth]{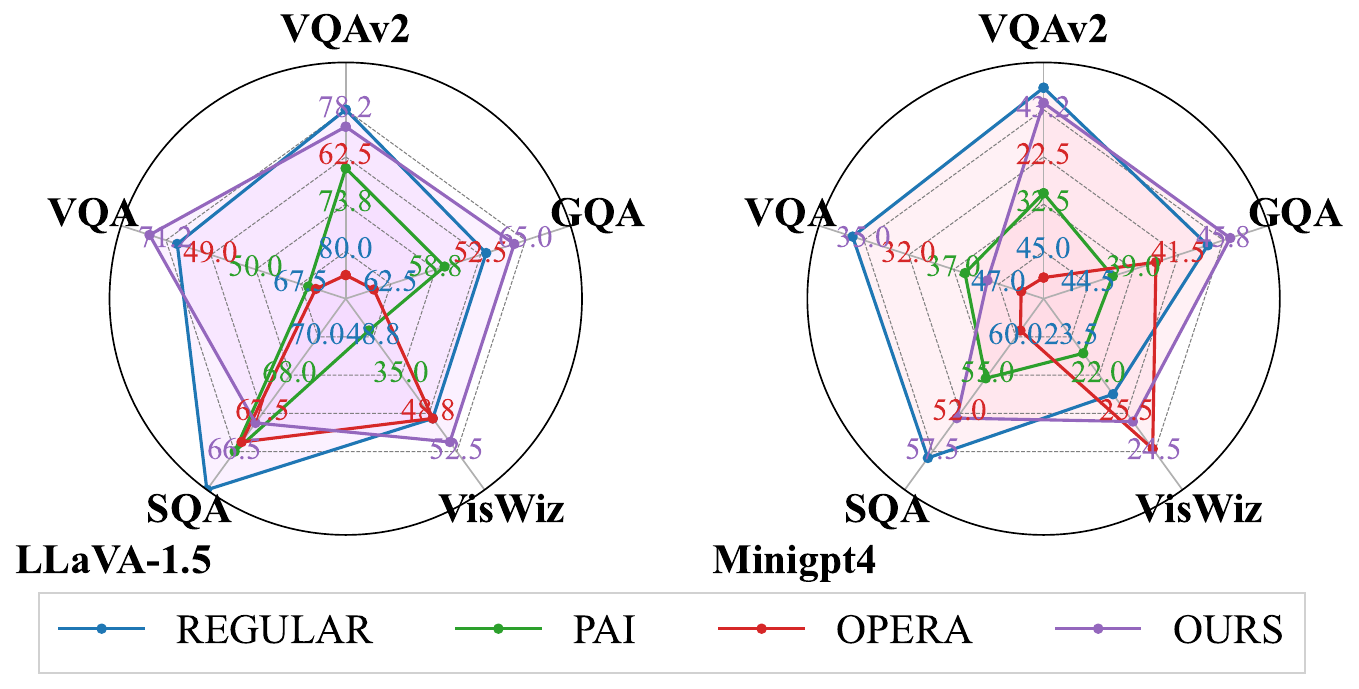}
\caption{
Comparison among different VLMs on five VQA benchmarks and three common benchmarks. 
The highest-performing results are highlighted in boldface.
}
\label{fig:VQA}
\vspace{-1em}
\end{figure}
\begin{figure}[t]
\centering
\includegraphics[width=1\linewidth]{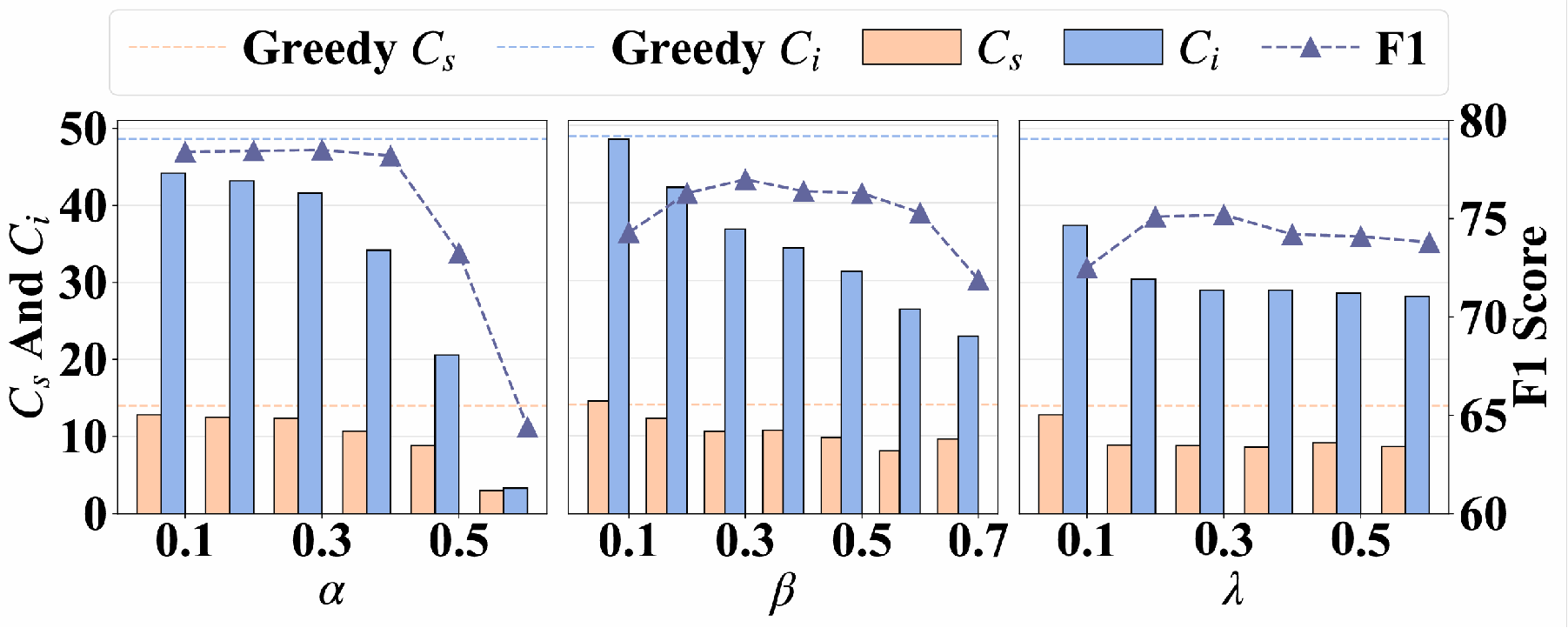}
\caption{
Impact of $\alpha$, $\beta$, and $\lambda$ on hallucination and informativeness in LLaVA-1.5, evaluated on $500$ COCO samples. 
}
\label{fig:ablation}
\end{figure}

\noindent\textbf{Results on vision-language understanding ability.}
To assess whether our hallucination mitigation hurts general vision-language understanding, we evaluate performance on five VQA benchmarks: VQAv2~\cite{vqa}, GQA~\cite{gqa}, VizWiz~\cite{vizwiz}, ScienceQA-IMG~\cite{science}, and TextVQA~\cite{towards}. 
As illustrated in Figure~\ref{fig:VQA}, \M~achieves consistent performance with the base LVLMs and even outperforms them on several benchmarks. 
For example, on LLaVA-1.5 compared to regular, we observe TextVQA (+3.7). Notably, gains on VizWiz (from $48.8$ to $52.5$, $7.6\%\uparrow$), where understanding visually degraded or text-heavy content is crucial. 
This suggests that our visual-attention-enhancing interventions help the model better localize and utilize visual cues under challenging conditions. 
Meanwhile, only a marginal decrease (from $80.0$ to $78.2$, $2.3$$\%\downarrow$) is observed on VQAv2, indicating minimal trade-offs in general capability.
Similar trends hold on MiniGPT-4, 
with a notable gain on GQA from $44.5$ to $45.8$ ($+1.3$), and a slight decrease on VQA-v2 from $45.0$ to $43.2$ ($-1.8$),
validating that our approach preserves or enhances reasoning under fine-grained or noisy visual contexts. 
Overall, these results confirm that our hallucination mitigation strategy maintains core vision-language understanding and even benefits tasks requiring robust visual grounding, thanks to our causal attention modulation and dual-path decoding.

\begin{figure*}[t]
\centering
\includegraphics[width=0.9\linewidth]{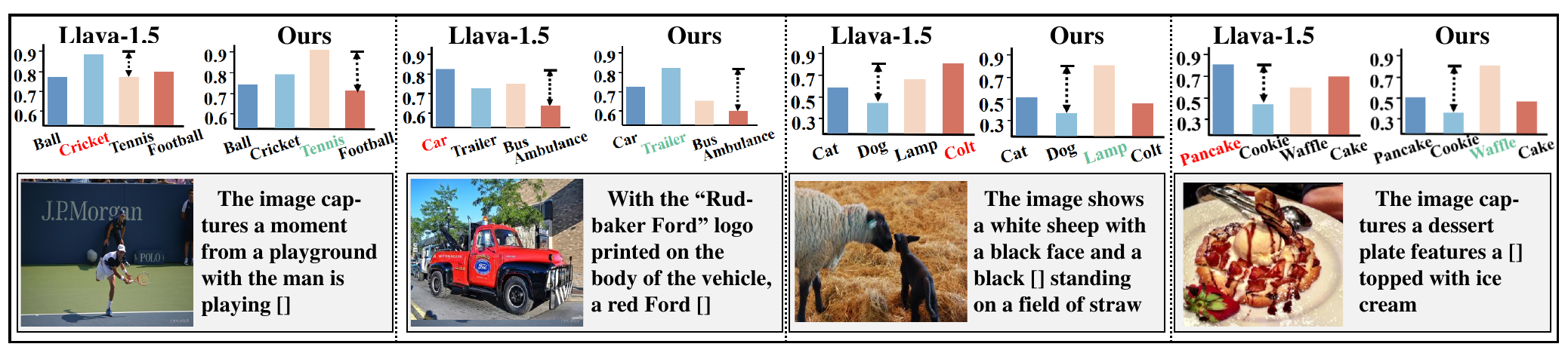}
\caption{
Visualization of token logits at hallucination-prone positions []. Red bars is hallucinated tokens and greens denote faithful ones. Our \M~consistently suppresses hallucinations and promotes visually-grounded predictions via DCD strategy.
}
\label{fig:logit}
\end{figure*}
\begin{figure}[t]
\centering
\includegraphics[width=0.9\linewidth]{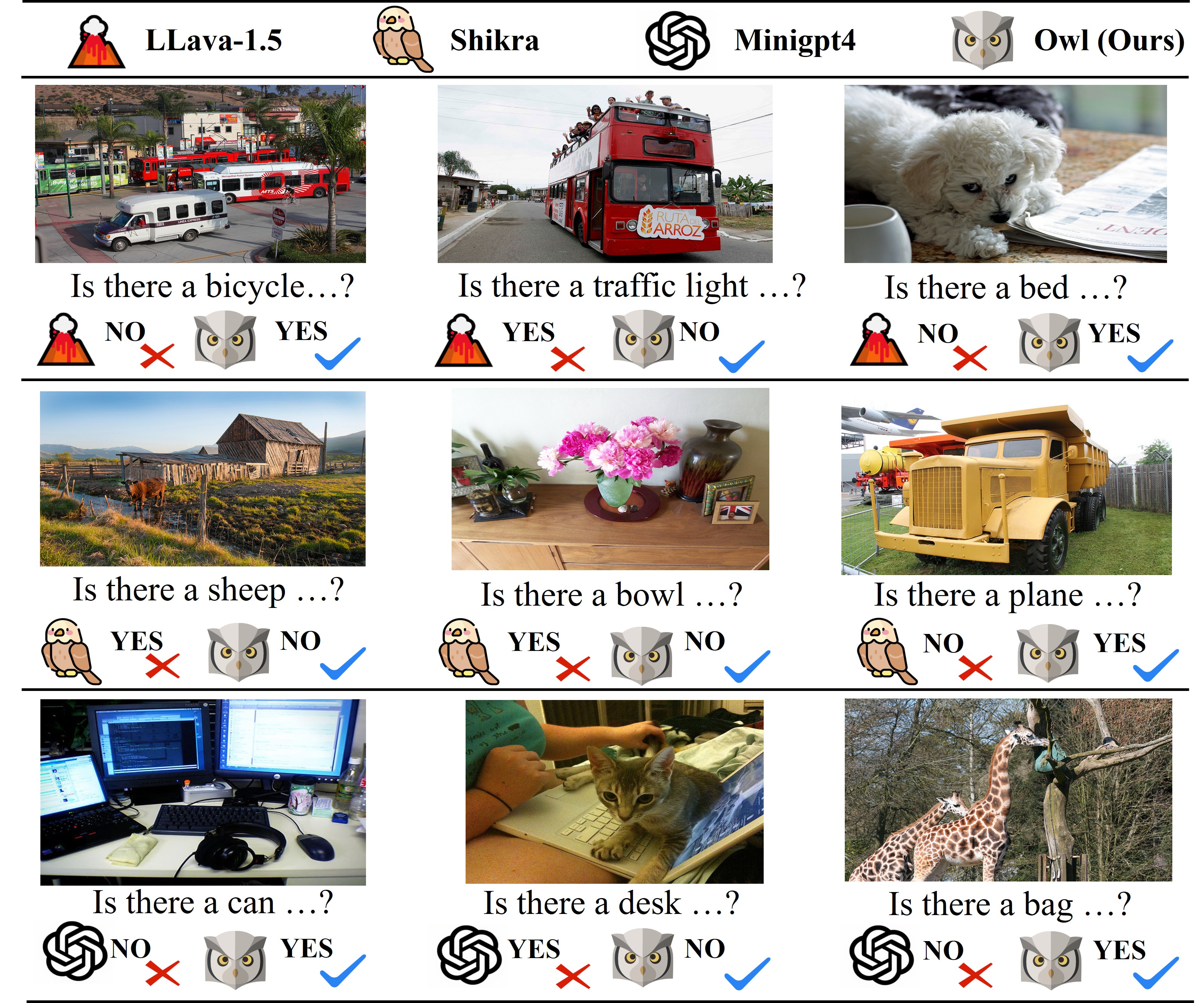}
\caption{Comparison on the POPE benchmark. Our \M~reliably avoids hallucinations and yields accurate predictions across diverse models and scenes.
}
\label{fig:pope}
\end{figure}

\paragraph{Ablation Study.}
We conduct ablations to assess the impact of three key hyperparameters in our DCD strategy: visual/textual attention coefficients $\alpha$, $\beta$, and the contrastive decoding strength $\lambda$. Evaluation is based on CHAIR for hallucination and F1 score for information richness and accuracy~\cite{liu2407paying}. As shown in the LLaVA-1.5 results in Figure~\ref{fig:ablation}, 
increasing $\alpha$ in Equations~\eqref{eqa:vp_alpha} and~\eqref{eqa:tp_alpha} enhances visual grounding and reduces hallucinations, but overly large values suppress informative content, lowering F1 score -- revealing a trade-off between hallucination mitigation and content richness. Increasing $\beta$ in Equations~\eqref{eqa:vp_beta} and~\eqref{eqa:tp_beta} steadily reduces hallucinations with a minimal drop in F1 score, suggesting that stricter regulation of textual attention effectively counters language priors without harming visual relevance, since the DCD strategy widens the gap between hallucination tokens and faithful tokens from Equation~\eqref{eq:CDC}. The parameter $\lambda$ balances visual and textual interventions during decoding. Moderate values ($0.1$--$0.4$) yield stable improvements, while excessively high $\lambda$ harms both CHAIR and F1, indicating decoding instability. In general, the results demonstrate that $\alpha$, $\beta$, and $\lambda$ play distinct but complementary roles. Careful tuning enables a balanced trade-off between hallucination suppression and semantic completeness.

\begin{figure}[t]
\centering
\includegraphics[width=0.9\linewidth]{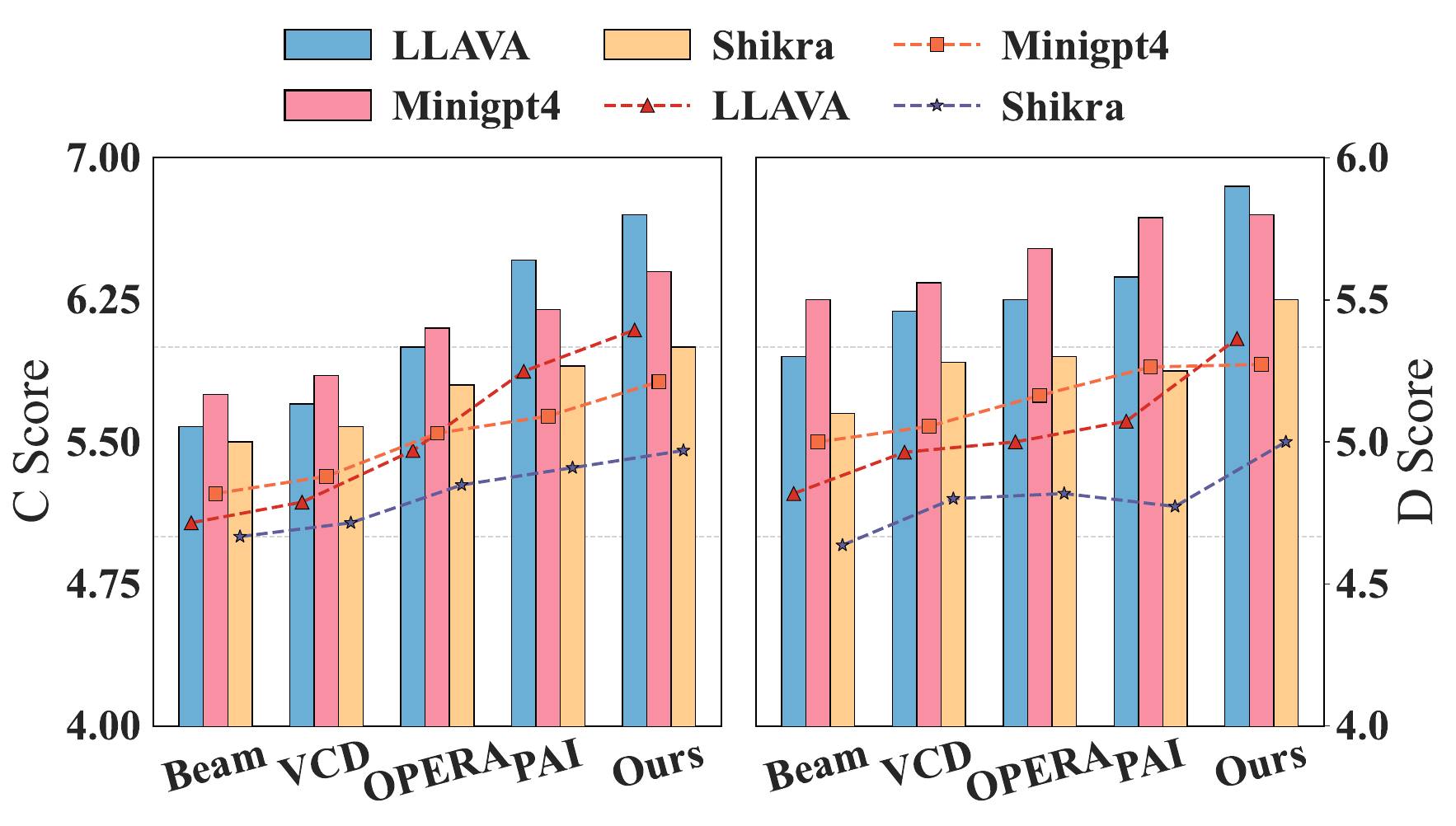}
\caption{
GPT-4V hallucination evaluation on MSCOCO. Left: Correctness (higher = less hallucination); Right: Detailedness. Line plots show model score trends per method.
}
\label{fig:gptv4}
\end{figure}

\noindent\paragraph{\textbf{Case study.}} 
To further illustrate the effectiveness of \M, we present two qualitative analyses.
In Figure~\ref{fig:logit}, we visualize the Top-$4$ token logits predicted by LLaVA-1.5 with and without our \M, which demonstrates how \M~consistently suppresses hallucinated tokens and promotes visually grounded predictions by adjusting attention distribution during decoding. 
Compared to LLaVA-1.5, which often assigns higher logits to misleading tokens influenced by language priors, our approach reweights token importance through visual enhancement and textual suppression, yielding more accurate results.
In Figure~\ref{fig:pope}, we visualize examples from the POPE benchmark, comparing \M~with three backbones. Although existing models frequently misidentify nonexistent objects (e.g. ``traffic light'', ``bowl,'' or ``bag''), \M~produces consistently correct responses. This highlights the strength of \M~framework in resisting hallucinations, particularly in cases where language priors strongly conflict with visual evidence.

\noindent\textbf{Results on human-like GPT-4v assisted hallucination evaluation.}
Figure~\ref{fig:gptv4} presents GPT-4V evaluations of \M~against beam search, VCD, OPERA, and PAI across three backbone models.
On LLaVA-1.5, our method lifts Correctness from $5.58$ to $6.70$ -- a notable improvement of $20.1\%$ -- and enhances Detailedness from $5.30$ to $5.90$ (up by $11.3\%$).
For MiniGPT-4, Correctness increases from $5.75$ to $6.40$ ($11.3\% \uparrow$), and Detailedness grows from $5.50$ to $5.80$ ($5.5\% \uparrow$).
Shikra shows a Correctness gain from $5.50$ to $6.00$ ($9.1\% \uparrow$), with Detailedness rising from $5.10$ to $5.50$ ($7.8\% \uparrow$).
Across all backbones, \M~consistently improves Correctness while maintaining or slightly enhancing Detailedness, with LLaVA-1.5 exhibiting the most significant gains.
These results confirm that our \M~framework effectively reduces hallucinations without compromising the richness of the generated content.

%% file: tables/CHAIR_result.tex
\begin{table}[b]
    \centering
    \setlength{\tabcolsep}{3pt}
    \resizebox{\linewidth}{!}{
    \begin{tabular}{c|ccc|ccc|ccc}
        \toprule
        \midrule
       \multirow{2}*{\textbf{Method}} & 
    \multicolumn{3}{c|}{\textbf{\emph{LLAVA-1.5}}} & 
    \multicolumn{3}{c|}{\textbf{\emph{MiniGPT-4}}} &
    \multicolumn{3}{c}{\textbf{\emph{Shikra}}}\\
         ~ & $C_S\downarrow$ & $C_I\downarrow$ & Len$\uparrow$ & $C_S\downarrow$ & $C_I\downarrow$ & Len$\uparrow$ & $C_S\downarrow$ & $C_I\downarrow$ & Len$\uparrow$ \\
        \midrule
    Beam Search & 48.6 & 16.2 & 105.3 & 33.0 & 8.7 & 77.8 & 53.8 & 17.4 & 116.2\\
    Greedy & 47.4 & 16.4 & 108.7 & 34.2 & 8.8 & 79.5 & 51.9 & 16.9 & 118.9\\
    Nucleus & 47.3 & 16.5 & 107.1 & 32.1 & 7.6 & 78.3 & 54.5 & 17.8 & 117.5\\
    \hline
    VCD & 44.6 & 14.4 & 93.8 & 33.1 & 11.2 & 71.2 & 47.2 & 15.5 & 105.6\\
    OPERA & 42.2 & 13.1 & 89.5 & 30.1 & 9.8 & 68.7 & 36.8 & 12.4 & 98.3\\
    PAI & 31.8 & 10.3 & 85.2 & 24.8 & 9.3 & 65.9 & 37.6 & 12.9 & 94.7\\
    CausalMM & 35.8 & 12.3 & 87.6 & 27.8 & 9.6 & 67.3 & 41.6 & 13.4 & 96.8\\
    \textbf{Ours} & \textbf{26.2} & \textbf{8.1} & \textbf{98.4} & \textbf{21.2} & \textbf{6.2} & \textbf{73.6} & \textbf{29.3} & \textbf{9.7} & \textbf{108.2}\\
    \midrule
        \bottomrule
    \end{tabular}
    }
    \caption{Results on CHAIR. $C_S$ and $C_I$ denote $\text{CHAIR}_S$ and $\text{CHAIR}_I$. 
    Len is the average length of generated text.
    }
    \label{tab:chair}
\end{table}

%% file: tables/POPE_results.tex
\begin{table}[t]
    \centering
    \setlength{\tabcolsep}{4pt}
    \resizebox{\linewidth}{!}{
    \begin{tabular}{c|ccc|ccc|ccc}
        \toprule
        \midrule
        \multirow{2}*{\textbf{Method}} & 
    \multicolumn{3}{c|}{\textbf{\emph{LLAVA-1.5}}} & 
    \multicolumn{3}{c|}{\textbf{\emph{MiniGPT-4}}} &
    \multicolumn{3}{c}{\textbf{\emph{Shikra}}} \\
    ~ & Ran$\uparrow$ & Pop$\uparrow$ & Adv$\uparrow$ & Ran$\uparrow$ & Pop$\uparrow$ & Adv$\uparrow$ & Ran$\uparrow$ & Pop$\uparrow$ & Adv$\uparrow$ \\
    \hline
    Beam Search & 84.6 & 84.4 & 83.1 & 69.2& 68.8& 67.4& 81.5 & 78.1 & 79.2 \\
    Greedy & 83.6 & 84.2 & 80.2 & 64.6& 63.3& 62.2& 81.9 & 78.1 & 80.2 \\
    Nucleus & 79.4 & 78.2 & 76.6 & 62.8& 59.8& 58.5& 80.2 & 76.5 &78.2 \\
    \hline
    VCD & 86.2 & 87.1 & 87.5 & 73.9& 71.2& 70.3& 81.2 & 77.3 & 80.8 \\
    OPERA & 87.6 & 88.2 & 90.1 & 75.8& 74.8& 72.4& 83.5 & 79.2 & 82.1 \\
    PAI & 89.8 &  \textbf{89.3} & 90.3 & 78.3& \textbf{79.8}& 78.9& 81.5 & 79.2 & 80.5\\
    CausalMM & 88.1 & 87.6 & 82.9 & 74.1& 73.6& 75.2& 83.4 & 79.6 & 81.7\\
    \textbf{\M (Ours)} & \textbf{90.2} & 88.1 & \textbf{90.5} & \textbf{82.2}& 78.4& \textbf{79.0}& \textbf{85.2} &\textbf{82.3} & \textbf{83.4} \\
    \midrule
        \bottomrule
    \end{tabular}
    }
    \caption{Results on POPE. Ran, Pop, and Adv is an abbreviation for \textit{Random}, \textit{Popular}, and \textit{Adversarial} setting, respectively. The higher score indicates better performance.}
    \label{tab:pope}
\end{table}

%% file: secs/7-conclusion.tex
\section{Conclusion}
In this work, we present a causally grounded framework \textbf{\M} for mitigating object hallucinations in LVLMs. By modeling the generation process through a structural causal model with visual and textual attention as mediators, we uncover the causal roots of hallucination and propose VTACR -- a novel metric to quantify the cross-modal attention balance during decoding. Guided by VTACR signals, we develop fine-grained attention interventions and a dual-path contrastive decoding strategy that effectively suppress hallucinated content. Extensive experiments on POPE and CHAIR benchmarks validate our approach, achieving a 22.9\% reduction in hallucination rates over strong baselines. 

This work provides both theoretical insights and practical tools for enhancing LVLM faithfulness, and opens new avenues for causal control in multimodal generation.

\section*{Acknowledgements}
This work was supported by the ``Gathering Resources to Revitalize Sichuan'' project initiated by the central government in Sichuan's higher education institutions (Grant No. 2025ZHCG0012), and Chengdu Achievement Transformation Demonstration Projec (Grant No. 2025YF0900067SN).